\documentclass[]{spie}  
\usepackage[]{graphicx}

\title{A Deep Drift-Diffusion Model for Image Aesthetic Score Distribution Prediction} 

\author{Xin Jin\supit{a} Xiqiao Li \supit{a} Heng Huang \supit{a} Xiaodong Li \supit{a}and Xinghui Zhou\supit{a}
\skiplinehalf
\supit{a}Beijing Electronic Science and Technology Institute; }

\authorinfo{Further author information: (Send correspondence to Xiaodong Li )\\Xiaodong Li : E-mail: lxdbesti@163.com}

\begin{document} 
  \maketitle 

\begin{abstract}
The task of aesthetic quality assessment is complicated due to its subjectivity. In recent years, the target representation of image aesthetic quality has changed from a one-dimensional binary classification label or numerical score to a multi-dimensional score distribution. According to current methods, the ground truth score distributions are straightforwardly  regressed. However, the subjectivity of aesthetics is not taken into account, that is to say, the psychological processes of human beings are not taken into consideration, which limits the performance of the task. In this paper, we propose a Deep Drift-Diffusion (DDD) model inspired by psychologists to predict aesthetic score distribution from images. The DDD model can describe the psychological process of aesthetic perception instead of traditional modeling of the results of assessment. We use deep convolution neural networks to regress the parameters of the drift-diffusion model. The experimental results in large scale aesthetic image datasets reveal that our novel DDD model is simple but efficient, which outperforms the state-of-the-art methods in aesthetic score distribution prediction. Besides, different psychological processes can also be predicted by our model.
\end{abstract}



\section{INTRODUCTION}
\begin{figure*}[htbp]
\centering
\includegraphics[width=\linewidth]{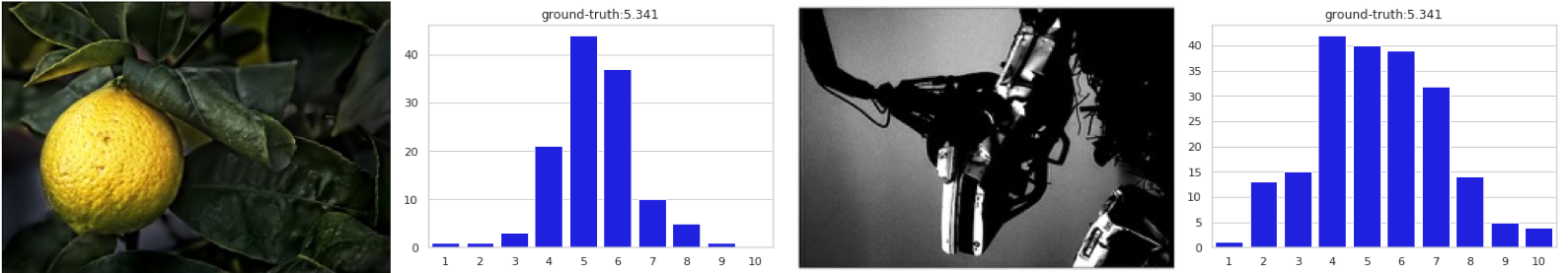}
\caption{Examples images and their aesthetic score distribution from AVA dataset. The aesthetic scores are rated from 1 to 10 and the vertical axis of the histogram represents the number of ratings.}
\label{fig:zhifangtu}
\end{figure*} 
Given the explosive growth of digital photography in the Internet and social networks, image aesthetic evaluation has received more and more attention due to its huge application potential. For example, current engines will retrieve and provide users with high aesthetic quality photos and guide aesthetic-driven image enhancement through aesthetic quality discriminators.Hence, it is desirable to automatically assess the aesthetic quality of the images.

Most previous works tried to use photography knowledge to guide the construction of artificial aesthetic features, such as the vivid colour, the rule of thirds, and the symmetry. However, these hand-craft and pre-defined features have limited representation ability, which is still a challenging task, although these features have shown encouraging results.

With the deep learning methods have shown great success in various computer vision tasks, more work has recently focused on using a deep convolutional neural network to extract effective aesthetics features. Image aesthetics assessment is typically cast as a classification or regression problem. Generally, the category or score of the image will be used as indicators to distinguish the level of aesthetic quality, and these approaches often combined some other information, such as object, scene and attribute, to guide the task of aesthetic assessment. However, people all have different ideas about what is beautiful. In other words, the process of giving an aesthetic evaluation of a picture may be quite different for different people. Two pictures of the same score label may have different evaluation processes. 

As Figure \ref{fig:zhifangtu},  shows six images and the associated aesthetic scores distribution  from the AVA database\cite{MurrayCVPR2012} . Each line has the same score, but the aesthetic distribution is different. Although the higher the aesthetic score, the more attractive. However, the use of a single label cannot effectively express the potential difference of human aesthetics. Therefore, using a single label is not enough to reflect the aesthetic psychology of the user.In contrast, aesthetics distribution prediction is a more reasonable way to evaluate the diversified aesthetics of images. Some works use aesthetic rating distributions as ground-truth, and then use various loss functions such as Kullback-Leibler(KL) divergence, cumulative Jensen-Shannon divergence\cite{JinAAAI2018} or earth movers distance\cite{NIMATIP18}.However, a most method is not considered the reason about the aesthetic distribution, more focus on reducing the distribution loss.
In this paper, we propose a Deep Drift-Diffusion (DDD) model inspired by psychologists to predict aesthetic score distribution from images. The DDD model simulates various positive and negative attractors and a disturbance factor based on the deep image features so that the psychological processes among user can be effectively expressed. The experimental results in public aesthetic image datasets (AVA \cite{MurrayCVPR2012} and Photo.net \cite{DattaECCV2006}) reveal that our novel DDD model outperforms the state-of-the-art methods on aesthetic score distribution prediction.

In summary, our main contributions are as follows:
\begin{enumerate}
\item The first work that embeds drift-diffusion psychological model into deep convolutional neural networks;
\item The combination of a dynamic model in psychology and score distribution prediction of visual aesthetics;
\item Our work has the potential of inspiring more attentions to model the psychology process of aesthetic perception beyond just modeling of the aesthetic assessment results.
\end{enumerate}

The main structure of this article is in the related work, we enumerate current work on aesthetic level assessment and aesthetic distribution evaluation. In the third chapter, we analyzed the statistical characteristics of the AVA dataset, and divided the data into various psychology model according to different math characteristics, and finally modeled different psychology models through deep learning. In Chapter 4, we compare the results of model with current methods for different aesthetic tasks. In Chapter 5, we summarized our work and put forward prospects for future work.

\section{Related Works}
\subsection{Image Aesthetics Assessment}
In the past decades, researchers on the topic of image aesthetic assessment have drawn much attention on 1D output: binary classification \cite{DattaECCV2006,luo2011content,kao2017deep,sheng2018attention} and aesthetic scoring \cite{KongECCV2016}. The binary classification is to give a binary label of image aesthetics: high or good and low or bad. The aesthetic scoring is to give a continues numerical score of image aesthetics: 0-1 or 0-10, the higher, the better.

However, only 1D binary label or aesthetic score can not fully describe the subjectiveness of aesthetic assessment. As claimed in Jin et al. \cite{JinAAAI2018},an image with similar scores may differ in the score distributions, which are histograms of scores given by multiple image reviewers. The scores are the mean of the score distributions. The other statistics of the distribution such as variations, skewness, kurtosis can be quite different from the images with similar scores.

In the past decades, image aesthetics assessment was mainly through human aesthetic perception of image features and photography rules. The features included spatial distribution of edges, color distribution, hue and blur etc \cite{luo2011content}. While drawing on some specific rules in photography, such as low depth-of-field indicator, then colorfulness measure, the shape convexity score and the familiarity measure\cite{DattaECCV2006}, The Rule of Thirds etc\cite{MurrayCVPR2012}, with the development of feature extraction technology, features based on high-level aesthetic principles had emerged, such as features based on scene types and related contents\cite{dhar2011high}, high-level semantic features based on this subject and background division\cite{luo2008photo}.

With the continuous advancement of deep learning research in recent years, CNN-based deep learning models were widely used in the classification and regression of aesthetics. Kao et al. proposed the multi-task learning\cite{kao2017deep}. They led the relevant items between tasks to the framework and made the utility of the appreciation of aesthetic and semantic labels more effective. Kong et al. and Chandakkar et al. utilized the relative aesthetics \cite{KongECCV2016} to select the new datasets of image from the datasets with the pairs of related labels, and trained the related image pairs to obtain the aesthetic ranking with the higher accuracy. Sheng et al. adopted attention-based multi-patch aggregation to adjust the weight of each patch in the training process. The results improved learning effectively \cite{sheng2018attention}.

\subsection{Score distribution prediction}
Recently, some methods are proposed to use modified or generated score distributions for binary classification and numerical assessment on aesthetics \cite{JinICIP2016,WangIJCNN2017,HouArXiv2016}. The pioneering
work of Wu et al. \cite{WuICCV2011} proposes a modified support vector regression algorithm to predict the score distribution in two small aesthetic datasets, before the large scale AVA dataset \cite{MurrayCVPR2012} released and the popularity of deep CNNs.

\begin{figure}[htbp]
\centering
\includegraphics[width=0.55\textwidth]{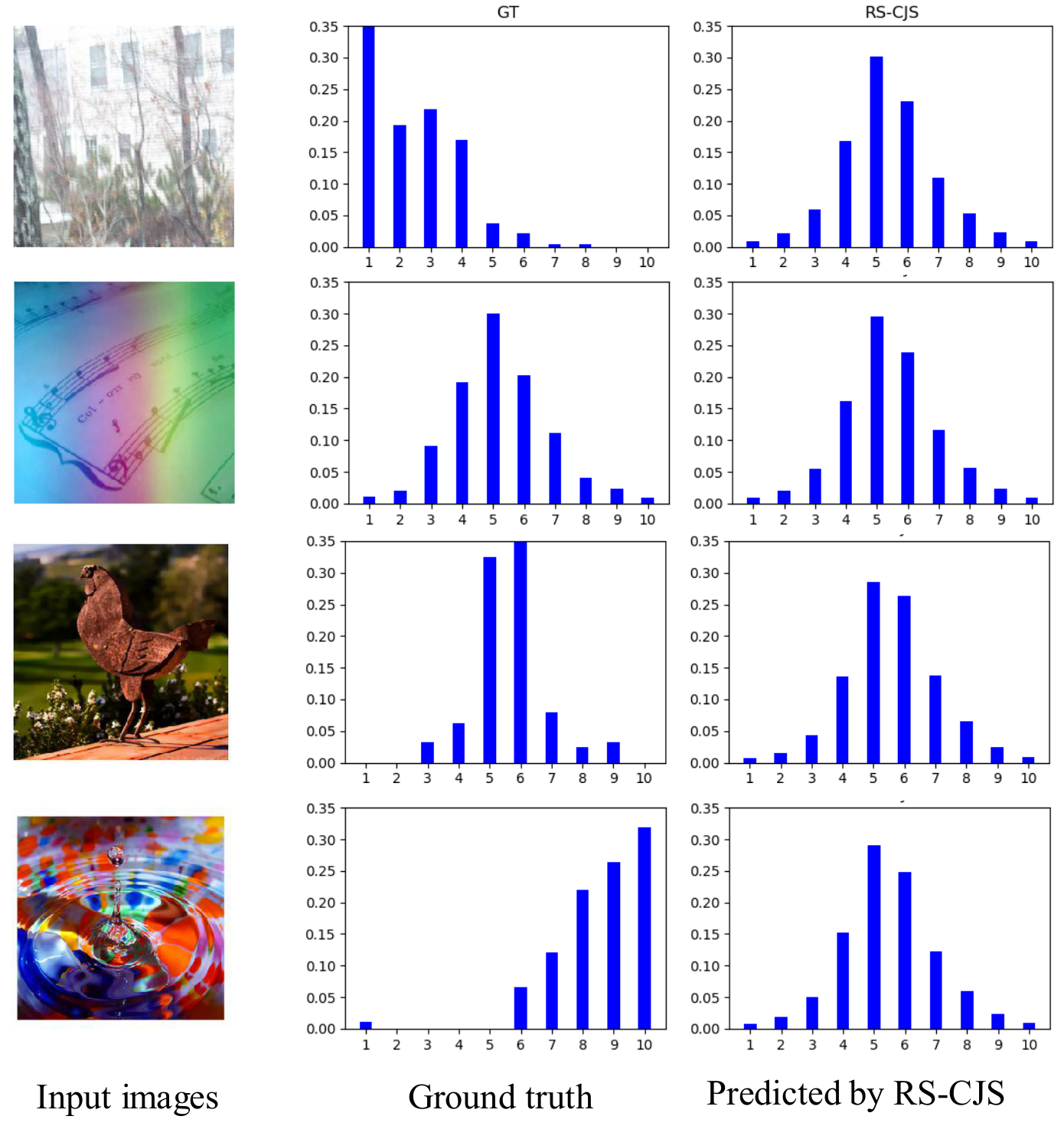}
\caption{Examples of non Gaussian distributions of AVA ratings with low, middle and high scores. The state-of-the-art method RS-CJS can predict Gaussian like distributions well but fails in the two ends: the low and the high non Gaussian distributions. Even some distributions in the middle are non Gaussian such as the one shown in the 3rd line.}
\label{fig:NonGaussian}
\end{figure} 

Most recently, Jin et al. propose a CNN based on the cumulative distribution with Jensen-Shannon divergence (RS-CJS) \cite{JinAAAI2018} to predict the aesthetic score distribution of human ratings, with a reliability-sensitive learning method based on the kurtosis of the score distribution. Talebi et al. propose a CNN based on EMD (Earth Mover’s Distance) loss \cite{NIMATIP18} to predict the aesthetic score distribution of human ratings. They use the predicted distribution to infer the mean score to guide the enhancement of images. Inspired by the human visual system, Xiaodan Zhang et al. propose the GPF-CNN architecture and GIF module\cite{zhang2019gated}. The former can learn to focus on the important
\begin{figure}[htbp]
\centering
\includegraphics[height=6.5cm]{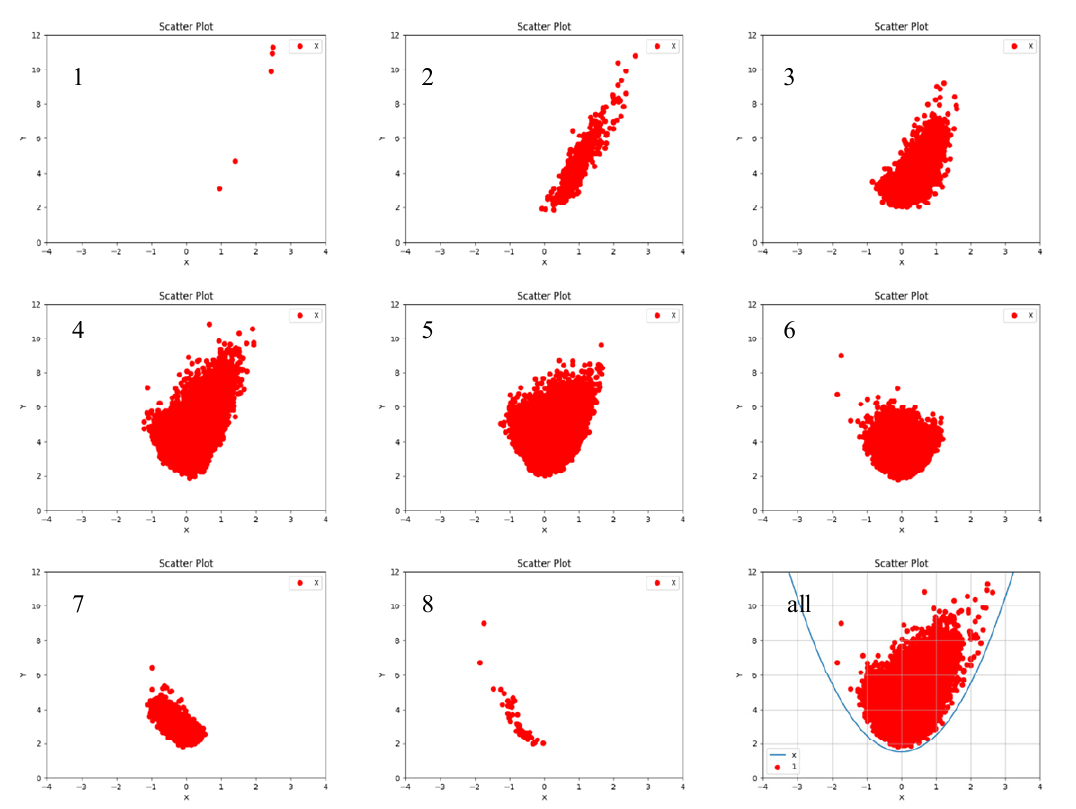}
\caption{The skewness-kurtosis maps of AVA score distributions. In each sub-plot, the horizontal axis and the vertical axis are the skewness and the kurtosis, respectively. Each point represent an images in AVA. The number in each sub-plot indicate the score section of images. In the last sub-plot, we plot all the points together.}
\label{fig:skmap}
\end{figure} 
 regions of the top-down neural attention map to extract fine detail features. The latter can adaptively fuse global and local features based on the input feature map. Gengyun Jia et al. consider that the aesthetic characteristics of images are a global feature \cite{jia2019theme}, and  resizing pictures during model training will affect the aesthetic characteristics. So they were applying ROI pooling on feature maps of their aesthetic distribution model. Qiuyu Chen et al.\cite{chen2020adaptive}  proposed an adaptive dilated convolution network to explicitly relate the aesthetic perception to the image aspect ratios while preserving the composition. Hui Zeng et al. create a comprehensive loss function to handle different aesthetic assessment tasks. Chaoran Cui et al. \cite{cui2018distribution} develop a novel deep neural model that can enhance image aesthetic evaluation by collect information from object classification and scene recognition.

\subsection{Subjectiveness of Aesthetics}
The human ratings are quite subjective \cite{KimTAC2018}. Unlike image recognition, people may give different scores of one image in the aspect of aesthetics. Chaoran Cui et al.\cite{cui2018distribution} also draw this conclusion from the entropy of the AVA \cite{MurrayCVPR2012}. The Gaussian distribution is the best-performing model for only 62\% of images in AVA. Examples of non-Gaussian distribution of human ratings are shown in Figure \ref{fig:NonGaussian}. The others are the skewed ones and can be best fitted by the Gamma distribution \cite{MurrayCVPR2012}.

The mean score is greatly influenced by the low and high extremes of the rating scale, which makes it inappropriate to be a robust estimation of the whole distribution, especially when the distribution is skewed. For skewed distributions, the median value appears to be more appropriate to describe the distributions than the mean value \cite{WuICCV2011}.

Before our work, many methods have achieved good performance on score distributions, such as method of \cite{JinAAAI2018,NIMATIP18,zhang2019gated,jia2019theme,cui2018distribution}. However, most of them follow the Gaussian's model. The mean score of the predicted distributions is always falling into [4:6].  This is because they have adopted a direct regression method.  62\% of the distributions approximately follow Gaussian, which leads the regression results to be as similar as Gaussian distributions \cite{JinAAAI2018}\cite{WangIJCNN2017}. Thus we need not only learn the results of human ratings, but also find the underline processes of human ratings. The math models such as Gaussian or Gamma can not well model the aesthetic score distribution. We should find a psychological model that describe the processes of aesthetic perception.

Tae-Suh Park et al. \cite{ParkTAC2015} make a thoroughly analysis of the consensus of visual aesthetic perception. They use the skewness-kurtosis maps (S-K map) to measure and visualise the consensus of aesthetic scores.The function is defined as $\mathrm{K}=\mathrm{S}^{2}+1$.  As shown in Figure \ref{fig:skmap}, the skewness and kurtosis of a score distribution are the high moments compared with mean and variance. With the assistance of S-K maps, they find four patterns of aesthetic score distributions from AVA dataset. None of Gaussian or Gamma distribution can model the four patterns, especially the wide range of kurtosis. They propose to use a dynamic psychological processes model from psychology to model the processes of aesthetic perception not only the aesthetic assessment results. 
In the discussion part of the work of Tae-Suh Park et al. \cite{ParkTAC2015}, they hope that the future work will combine a dynamic model in psychology and score distribution prediction of visual aesthetics.

\begin{figure}[!htbp]
\centering
\includegraphics[width=0.43\textwidth]{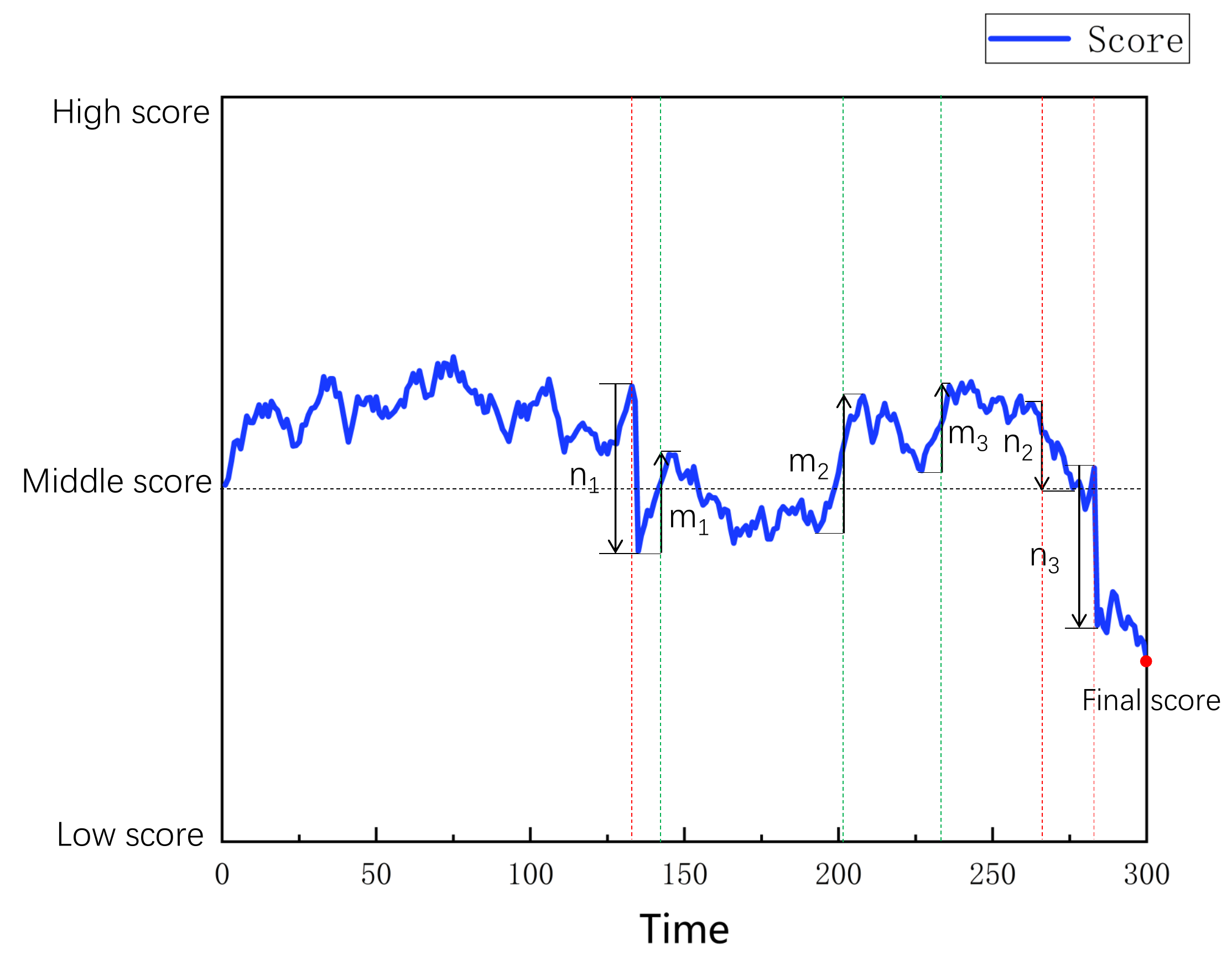}
\caption{The simulation of the psychological processes of aesthetic perception of one rater. From the middle score (5 in AVA), the score will be disturbed by white noise. When a positive factor attracts the rater, the score will increase according to a exponential distribution, and vice verse. In this example, there are 3 positive attractors and 3 negative attractors. The final score is about 2.5.}
\label{fig:process}
\end{figure}
\begin{figure*}[p]
\centering
\includegraphics[height=22cm]{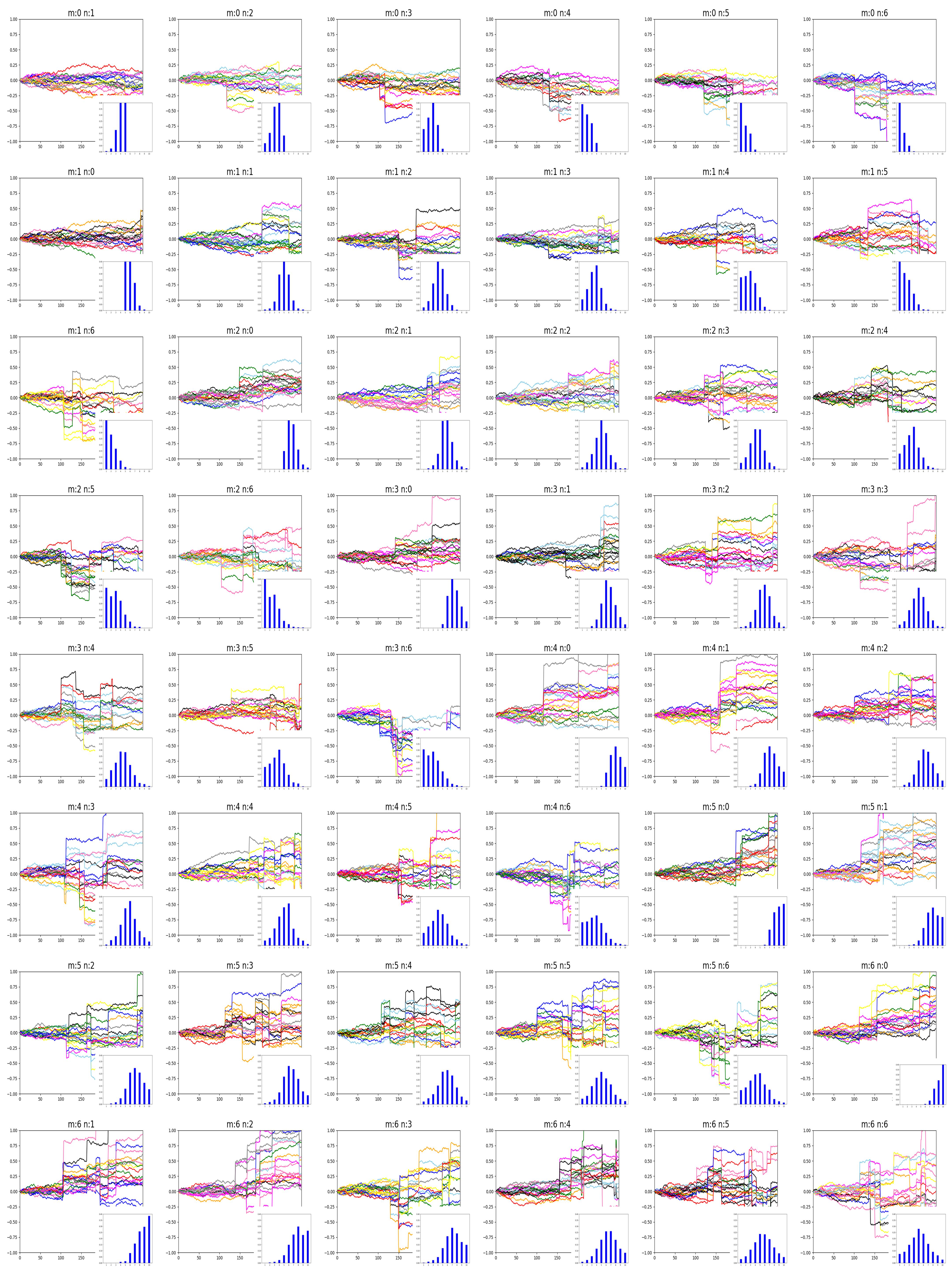}
\caption{Score distributions simulated by n, m. $n,m \in [0,6]$}
\label{fig:simulate20}
\end{figure*}

\section{The Proposed DDD Model}
\subsection{Subjectiveness Analysis}
Jin et al. \cite{JinAAAI2018} make a statistical
analysis of subjectiveness or diversity of the opinion among
annotators in a large-scale database for aesthetic visual analysis
(AVA) \cite{MurrayCVPR2012}. This dataset is specifically constructed for the purpose of learning more about image aesthetics. All those images are directly downloaded from dpchallenge.com. For each image in AVA, there is an associated distribution of scores (1-10) voted by different viewers. The number of votes that per image gets is ranged in 78-549, with an average of 210.

It is observed that most images’ mean values are located in [4:7]. Images in this interval are not easy to be classified to a high-low label. Most images’ standard deviation values are larger than 1:25, which shows the diversity of the human ratings for the same image. Images with mean score values from four to seven tend to have a low absolute value of the skewness and can be considered as those with symmetrical score distributions. Images with mean score values lower than four and greater than seven can be considered as those with positively and negatively skewed score distributions, respectively. This is likely due to the non-Gaussian nature of score distributions at the extremes of the rating scale. Within each range of the mean scores, there exist some images with high absolute values of kurtosis values (after normalized by minus 3), which are considered as those with unreliable score distributions \cite{JinAAAI2018}. Their exist wide range of kurtosis. This is the core reason that Gaussian distribution can not well fit the human ratings.

\subsection{The Deep Drift-Diffusion Model}
Inspired by Tae-Suh Park et al. \cite{ParkTAC2015}, we propose a Deep Drift-Diffusion (DDD) model based on psychology to predict the aesthetic score distribution of images. The DDD model combines the deep convolutional neural networks of artificial intelligence and the dynamic drift model from psychology \cite{KelsoLNCS1995,RatcliffPsyRev1978,BogaczPsyRev2006,RatcliffNC2008} The DDD model simulates various positive and negative attractors and a disturbance factor based on the deep image features.

As shown in Figure \ref{fig:process}, A person's aesthetic view is regarded as a process of being attracted by positive factors and influenced by negative factors. Suppose the initial score of a figure is 5 (5 in AVA),then the score will be disturbed by white noise. When a positive factor attracts the evaluators, the score will increase according to an exponential distribution. When the evaluator is affected by negative factors, the score will decrease according to an exponential distribution. The drift-diffusion model is described in Eq. \ref{eq:DDD}.

The AVA dataset is from an online website. The photo reviews are given sufficient time to find the advantages and disadvantages of the aesthetics. Thus, the time factor does not that affect the results. Although there is no temporal component to the decision process in AVA dataset, the time factor in the process of psychological evaluation does not directly affect the outcome of the evaluation.

Where $E_{pos}$ and $E_{neg}$ follow exponential distribution (Eq. \ref{eq:exponential}). W is our modified white noise as shown in Eq. \ref{eq:whitenoise}. U is a uniform random distribution. Positive attractors represent the fluctuations of  the score caused by good aesthetic factors, which can be simply interpreted as the advantages of the picture discovered by the viewer. While negative attractors are the opposite. The computational methods of $E_{pos}$ and $E_{neg}$ are consistent. Since they represent the same psychological status in the model, they must be characterised by the same distribution. According to this formula, one person's score is represented at a time of running. The score distribution can be recalculated by calculating the result of the formula for the number of times in the AVA images, thus generating training and testing labels. One example of 20 raters on one image with various ratios of m:n are shown in Figure \ref{fig:simulate20}.

\begin{equation}
v = MiddleScore + \sum_{i = 1}^{m}E_{pos} - \sum_{i = 1}^{n}E_{neg} + W,
\label{eq:DDD}
\end{equation}

\begin{figure}[htbp]
\centering
\includegraphics[height=0.8\linewidth]{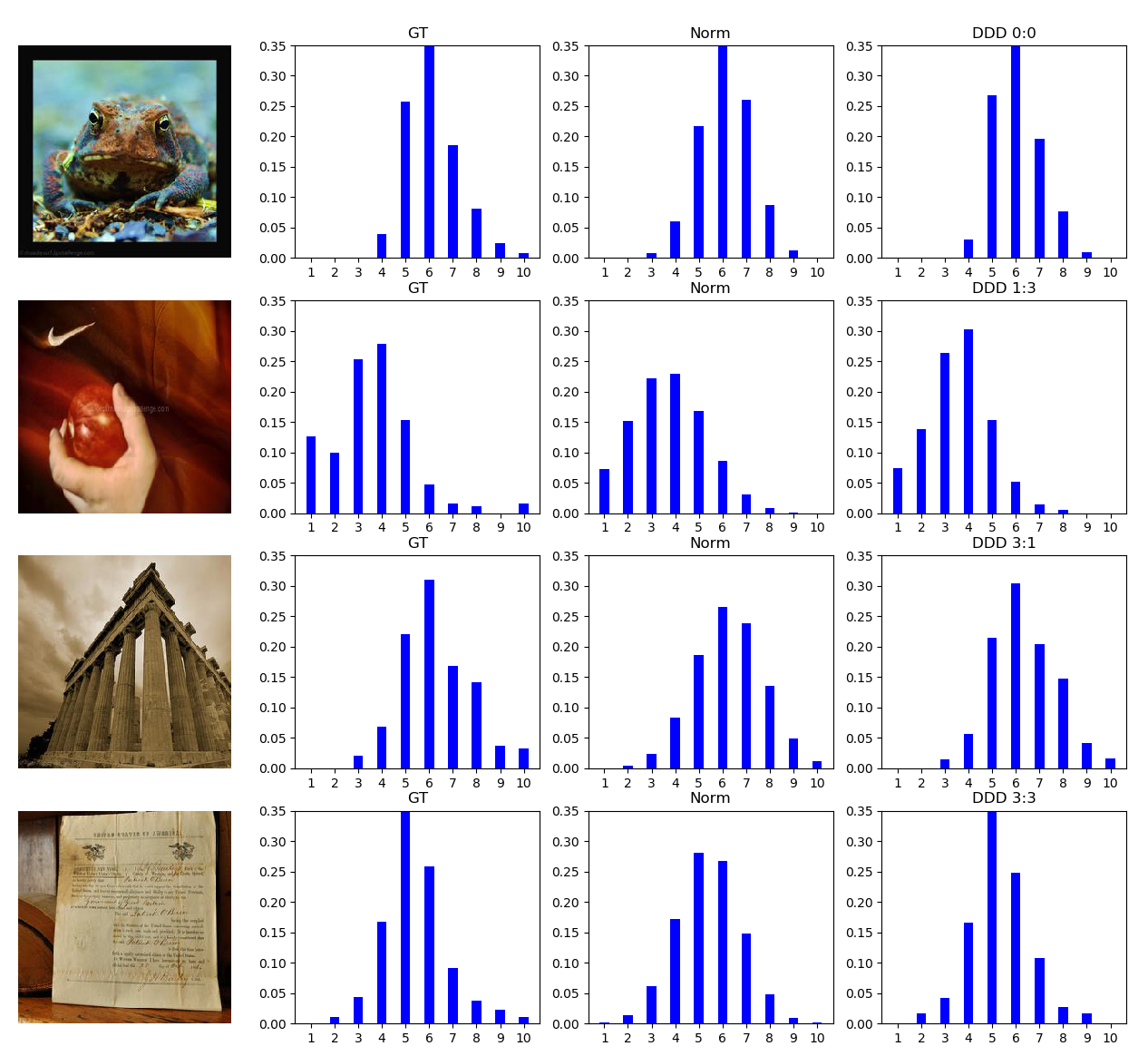}
\caption{Examples of the fitting using Gaussian and our DDD model. Obviously, in the two extreme ends, the results of our DDD model are more similar as ground truth values. Images are from AVA dataset.}
\label{fig:GaussianVSDDD}
\end{figure}

\begin{equation}
E = 0.5 * e^{-0.5 * U(0,10)}
\label{eq:exponential}
\end{equation}

\begin{equation}
W = 0.015 * U(-1, 1),
\label{eq:whitenoise}
\end{equation}
where the parameter 0.015 is from \cite{ParkTAC2015}.


We fit the ground truth score distributions using the Gaussian model and our DDD model respectively. The number of positive and negative attractors are determined by an exhaustive search. The ones who get the smallest distance between the simulated and the ground truth distributions are selected. The upper bound of the number of the positive or negative abstractors is 7, which is determined by experiments. The value above 7 does not decrease the fitting errors any more. Besides, In order to accurately calculate the exact positive and negative attractors, we enumerated the combination of all positive and negative attractors. Thus, we simulate 49 categories of psychological processes.

Through the combination of the number of positive and negative attractors, we can form 49 different specific quantitative relationships, which can represent 49 different psychological processes. And Tae-Suh Park and otherts\cite{ParkTAC2015} have only 4 psychological processes. Four examples of the fitted results are shown in Figure .\ref{fig:GaussianVSDDD}. Besides, we make numerical evaluation of the fitted results.

As shown in Table \ref{tb:fit}, in the middle, the Gaussian model wins our DDD a little. While at the two extreme ends, the DDD model can fit better than Gaussian model does. In addition, the DDD fits the four moments of the aesthetic distributions much better than the Gaussian does, as shown in Table \ref{tb:moments}.

\begin{figure*}[!htbp]
\centering
\includegraphics[width=0.8\linewidth]{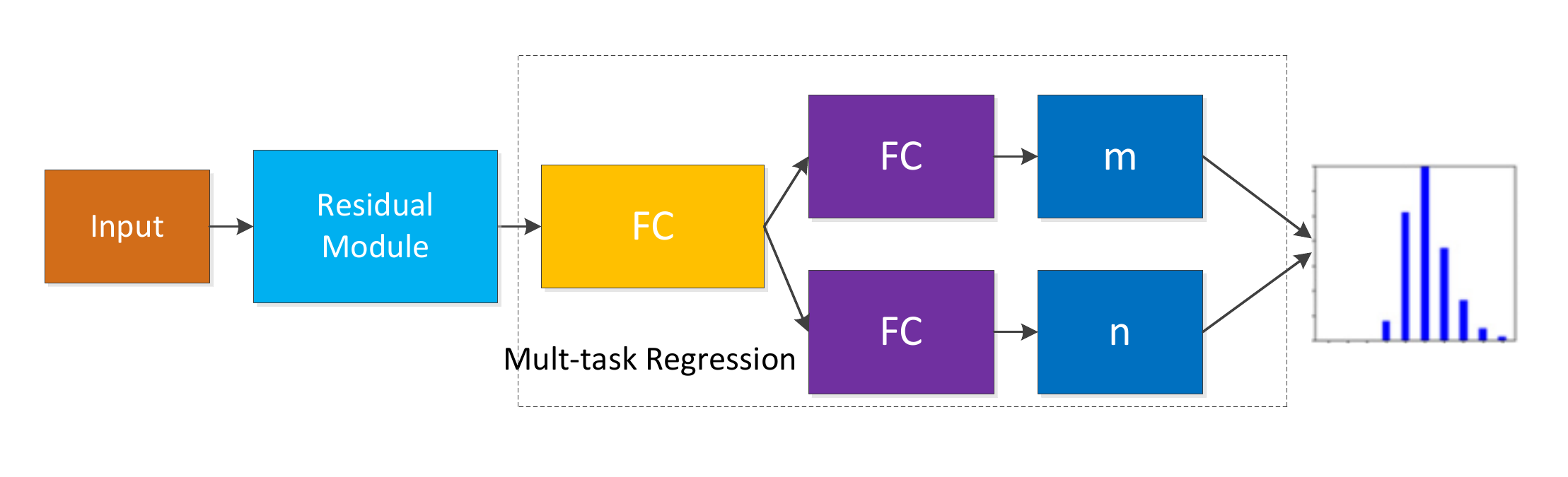}
\caption{The architecture of the proposed deep drift-diffusion model. The ResNet-50 is our baseline CNN.}
\label{fig:DDD}
\end{figure*}
 
The architecture of our DDD model is shown in Figure \ref{fig:DDD}. We use a ResNet-50 as our basic model. Then we attach a multi-task regression module which contains the number of the positive ($m$) and the negative ($n$) attractors. The fitted model parameters are used as the ground truth labels for training. However, when evaluating our models in the experiments, we compare the distributions generated by our predicted parameters $m$ and $n$ with the original ground truth score distributions.

\begin{figure*}[tbp]
\centering
\scalebox{0.85}{\includegraphics{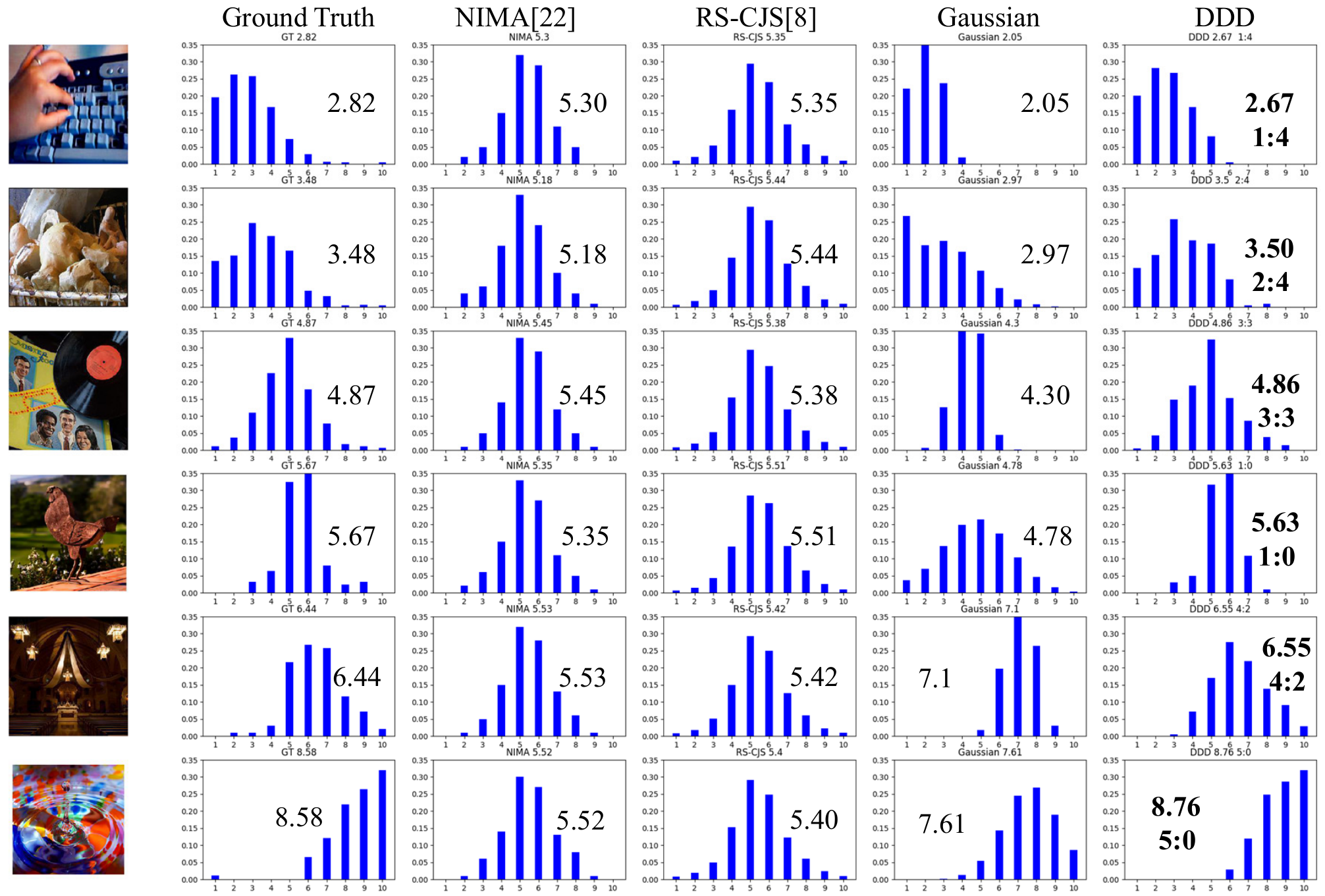}} 
\caption{The comparison of aesthetic distribution prediction. The network used in this figure is ResNet-50. Images are from AVA dataset. Note that, different psychological processes can also be predicted by our DDD model with different ratios of $m$ and $n$.}
\label{fig:Distribution}
\end{figure*}

\section{Experiments}
\subsection{Datasets}


To the best of our knowledge, the AVA dataset and the Photo.net \cite{DattaECCV2006} dataset is the only two publicly available datasets with ground truth aesthetic score distributions. These two datasets are popular in the computational aesthetic community. We evaluate our DDD model on both AVA dataset \cite{MurrayCVPR2012} and Photo.net dataset \cite{DattaECCV2006}. We compare  our models with the state-of-the-art methods of score distribution prediction task. Our DDD model outperforms all the state-of-the-art methods.

\begin{table*}
\caption{The fitting errors by Gaussian and DDD, we use RMSE (Rooted Mean Square Error) to evaluate the errors.}
\label{tb:fit}       
\centering
\begin{tabular}{lllllllllll}
\hline\noalign{\smallskip}
Methods & 1-2 & 2-3 & 3-4 & 4-5 & 5-6 & 6-7 & 7-8 & 8-9 & all \\
\noalign{\smallskip}\hline\noalign{\smallskip}
Gaussian & 0.117 & 0.063 & 0.030 & \textbf{0.309} & \textbf{0.305} & \textbf{0.274} & 0.276 & 0.524 & 0.0302 \\
\textbf{DDD Model} & \textbf{0.026} & \textbf{0.030} & 0.030 & 0.311 & 0.309 & 0.275 & \textbf{0.265} & \textbf{0.310} & \textbf{0.0294} \\
\noalign{\smallskip}\hline
\end{tabular}
\end{table*}

\begin{table*}
\caption{The fitting errors of four moments by Gaussian and DDD, we use MSE ( Mean Square Error) to evaluate the errors.}
\label{tb:moments}       
\centering
\begin{tabular}{llllll}
\hline\noalign{\smallskip}
Methods & $\mu$(MSE) & $\sigma$(MSE) & skew(MSE) & kurt(MSE) \\
\noalign{\smallskip}\hline\noalign{\smallskip}
Gaussian & $\approx$0 & 0.0001 & 0.1951 & 1.0391 \\
\textbf{DDD Model}  & $\approx$0 & 0.0001 & \textbf{0.0805} & \textbf{0.4725} \\
\noalign{\smallskip}\hline
\end{tabular}
\end{table*}

\begin{table*}

\caption{The comparison of aesthetic distribution  prediction on AVA dataset. The lower the better.}
\label{tb:Distribution}       
\centering
\begin{center}  

\begin{tabular}{lcccccccccc}
\hline\noalign{\smallskip}
Methods & PED & PCE & PJS & CED & CJS & PCS & PKL  & EMD & Class. Acc. $\uparrow$ \\
\noalign{\smallskip}\hline\noalign{\smallskip}

Gaussian(ResNet-101)& 0.162  & 2.817 &  0.048&0.254  & 0.050 & 0.076& 0.381  &-&- \\
RS-CJS(1/3GoogleNet) \cite{JinAAAI2018} & 0.158  & 2.760 &  0.037&0.260  & 0.040 & 0.068& 0.323 &-&80.08\% \\
NIMA(inception-v2) \cite{NIMATIP18}& 0.168 & 2.693 & 0.028 & 0.137 &0.029  & 0.044 & 0.081  & 0.050&81.51\% \\
Hui Zeng et al. \cite{zeng2019unified} & - & -&-  &-  &-  &-  &  0.101 & 0.065&80.81\% \\
Chaoran Cui et al. \cite{cui2018distribution}& 0.127 &  -& - &- & - & -  & 0.094 &- & -\\
Gengyun Jia et al.\cite{jia2019theme} & - & - & - &  -& - & - & -& 0.041&- \\
Xiaodan Zhang et al.\cite{zhang2019gated} &  -&  -& - & -&  -&  -&  - &0.045&81.81\%\\

\hline
DDD(1/3GoogLeNet) & 0.142 & 2.729 & 0.028 & 0.177 & 0.026 & 0.051 & 0.153  &0.044  &81.59\%\\
DDD(MobileNet-v1) & 0.125 & 2.669 & 0.022 & 0.138 & 0.019 & 0.042 & 0.092 & 0.031&80.43\% \\
DDD(ResNet-50) & 0.109 & 2.667 & 0.020 & 0.129 & 0.015 & 0.035 & 0.071 &0.026   &82.63\%\\
\textbf{DDD(ResNet-101)} & \textbf{0.105} & \textbf{2.640} & \textbf{0.019} & \textbf{0.122} & \textbf{0.013} & \textbf{0.028} & \textbf{0.065}& \textbf{0.023}& \textbf{82.65\%}\\
\noalign{\smallskip}\hline
\end{tabular}
\end{center}
\end{table*}

\textbf{AVA}. The AVA dataset is a list of image ids from DPChallenge.com, which is an online photography social network. There are total 255,530 photographs, each of which is rated by 78–549 persons, with an average of 210 aesthetic ratings ranging from 1 to 10. We follow the standard partition method of the AVA dataset in previous work
\cite{MurrayCVPR2012,WangSP2016,KongECCV2016,LuTMM2015,MaiCVPR2016,JinAAAI2018}. The training and testing sets contain 235,599 and 19,930 images respectively.

\textbf{Photo.net}.  Each image in the Photo.net dataset is rated by at least ten users to evaluate the aesthetic quality from 1 to 7. Due to some unavailable links in photo.net website, we collect 15,582 images in all. We follow the partition ratio in previous work \cite{DattaECCV2006,KaoTIP2017}. The training and testing sets contain 13,582 and 2000 images, respectively. For the aesthetic quality classification task, we also follow \cite{DattaECCV2006,KaoTIP2017} and choose the average score of $5.0$ as median aesthetic ratings. The images with an average score larger than $5+\delta$ are designated as high quality images, those with an average score smaller than $5$ as low-quality images. We set $\delta$ to $0$ in the experiment, which is more challenging than that with setting $\delta$ to other values \cite{MurrayCVPR2012}.

\begin{table*}
\caption{The comparison of aesthetic distribution prediction on Photo.net dataset. The lower the better.}
\label{tb:Distribution_photonet}       
\centering
\begin{tabular}{lllllllllll}
\hline\noalign{\smallskip}
Methods & PED & PCE & PJS & CED & CJS & PCS & PKL & EMD & Class. Acc. $\uparrow$ \\
\noalign{\smallskip}\hline\noalign{\smallskip}
Gaussian(1/3GoogLeNet) & 0.313 & 2.351 & 0.097 & 0.348 & 0.066 & 0.179 & 1.432 & 0.075&73.54\%\\
RS-CJS(1/3GoogLeNet)\cite{JinAAAI2018} & 0.305 & 2.270 & 0.085 & 0.311 & 0.060 & 0.143 & 1.340&0.072&75.62\%\\
DDD(1/3GoogLeNet) & 0.289 & 2.208 & 0.073 & 0.260 & 0.054 & 0.121 & 1.247&0.070&77.96\%\\
\hline
Gaussian(ResNet-50) & 0.296 & 2.164 & 0.093 & 0.292 & 0.064 & 0.153 & 1.273&0.073&76.76\%\\
RS-CJS(ResNet-50)\cite{JinAAAI2018}  & 0.262 & 1.963 & 0.071 & 0.264 & 0.059 & 0.138 & 1.185&0.069&78.10\%\\
DDD(ResNet-50) & 0.243 & 1.842 & 0.068 & 0.251 & 0.106 & 0.035 & 1.129&0.066&79.22\%\\
\hline
Gaussian(ResNet-101) & 0.293 & 2.157 & 0.092 & 0.289 & 0.061 & 0.149 & 1.268&0.071&76.84\%\\
RS-CJS(ResNet-101) \cite{JinAAAI2018} & 0.255 & 1.961 & 0.070 & 0.262 & 0.058 & 0.137 & 1.183&0.068&78.12\%\\
\textbf{DDD(ResNet-101)} & \textbf{0.242} & \textbf{1.840} & \textbf{0.068} & \textbf{0.249} & \textbf{0.047} & \textbf{0.105} & \textbf{1.126}& \textbf{0.064}& \textbf{79.26\%}\\
\noalign{\smallskip}\hline
\end{tabular}
\end{table*}

\subsection{Implementation Details}
We fix the parameters of the layers before the first fully connected
layer of a pre-trained GoogLeNet model and ResNet mode1 \cite{HeCVPR2016} on the ImageNet  and fine-tune the all full connected
 layers on the training set of the AVA dataset. We use the
Caffe framework to train and test our models.
The learning policy is set to step. Stochastic gradient descent
is used to train our model with a mini-batch size of 48 
images, a momentum of 0.9, a gamma of 0.5 and a weight
 decay of 0.0005. The max number of iterations is 120000. 
The training time is about 5 hours and 8 hours using Titan X Pascal GPU.

\subsection{Score Distribution Prediction}
We compare our DDD model with the method of RS-CJS \cite{JinAAAI2018}, which uses 1/3 GoogLeNet and the RS-CJS loss. The evaluation rules follow those in \cite{JinAAAI2018}. The numerical results are shown in Table \ref{tb:Distribution} (AVA) and Table \ref{tb:Distribution_photonet} (Photo.net). For a fair comparison, we also use 1/3 GoogLeNet to replace the ResNet-50 in Figure \ref{fig:DDD}. Besides, we modify the regression targets of our DDD to the $\mu$ and $\sigma$ of the fitted Gaussian distributions. The numerical results in Table \ref{tb:Distribution} and Table \ref{tb:Distribution_photonet} reveal that our DDD model beats the Gaussian model and the RS-CJS, which directly fit the results no matter using 1/3 GoogLeNet, ResNet-50 or ResNet-101. The performances of the Gaussian model are even worse than RS-CJS \cite{JinAAAI2018}.

We also give some visualized comparison results in Figure \ref{fig:Distribution}. Scores of most images of AVA dataset are in the range of [4,6]. Thus the NIMA \cite{NIMATIP18} and RS-CJS \cite{JinAAAI2018} tend to output distributions in the middle. The mean scores of their distributions also fall in [4,6]. The regression errors are less influenced by the image with very low or very high scores. The Gaussian distribution can not fit the original distribution well. The predicted distributions of our DDD model fit well the ground truth distributions. We can not only get output middle scores but also low and high scores. Besides, the images with middle scores can be divided by the different ratios of $m$ and $n$ of the DDD model, which represent different kinds of psychological processes.

As shown in Table \ref{tb:Distribution} and Table \ref{tb:Distribution_photonet}, our DDD model outperforms all the state-of-the-art methods \cite{JinAAAI2018,NIMATIP18,zeng2019unified,zhang2019gated,jia2019theme,cui2018distribution}  on the aesthetic score distribution prediction task using the evaluation metrics of \cite{JinAAAI2018}, which contain 8 different kinds of distribution distances. All the previous state-of-the-art methods do not take the psychological process of aesthetics into consideration. They only model the aesthetic perception results in the form of score distributions of multiple reviewers.

\subsection{Aesthetic Classification}
We recast our predicted score distribution to 1-dimensional binary label and compare with the state-of-the-art methods on aesthetic quality classification, as shown in the last column of Table \ref{tb:Distribution}. Compared to the state-of-the-art method on aesthetic score distribution prediction \cite{JinAAAI2018,NIMATIP18,zhang2019gated,zeng2019unified}, our DDD model achieves the state-of-the-art aesthetic classification on AVA dataset. Note that, on the AVA dataset, our DDD model with only 1/3 GoogleNet beats NIMA model \cite{NIMATIP18} with full Inception-v2 GoogleNet on the aesthetic classification task.


\section{Conclusions and Discussions}
In this paper, we propose a DDD model inspired by psychologists to predict aesthetic score distribution from images. The DDD model simulates various positive and negative attractors and a disturbance factor based on the deep image features. The experimental results in large scale aesthetic image datasets (AVA and Photo.net) reveal that our novel DDD model outperforms the state-of-the-art methods in aesthetic score distribution prediction. Besides, different psychological processes can also be predicted by our model. 

The original drift diffusion psychology model has a temporal component of the decision process. In future work, we will build such an aesthetic dataset with score distributions and rating time data, which models the dynamic process of the aesthetic perception. Besides, we will explore more psychological processes of aesthetic perception.

\bibliography{article}   
\bibliographystyle{spiebib}   

\end{document}